\documentclass[runningheads]{llncs}
\usepackage{graphicx, amsmath, bbm, multirow, subcaption}
\usepackage[table]{xcolor}
\usepackage{colortbl}
\begin{document}
\title{Which Client is Reliable?: A Reliable and Personalized Prompt-based Federated Learning for Medical Image Question Answering}
%
%
\author{He Zhu$^{\dag}$, Ren Togo$^{\dag\dag}$, Takahiro Ogawa$^{\dag\dag}$ and Miki Haseyama$^{\dag\dag}$}

%
%
\institute{    
    $^{\dag}$Graduate School of Information Science and Technology, Hokkaido University, Japan\\
    $^{\dag\dag}$Faculty of Information Science and Technology, Hokkaido University, Japan\\
    E-mail: \{zhu, togo, ogawa, mhaseyama\}@lmd.ist.hokudai.ac.jp
    }
%
\maketitle              
\begin{abstract}
Conventional medical artificial intelligence (AI) models face barriers in clinical application and ethical issues owing to their inability to handle the privacy-sensitive characteristics of medical data. We present a novel personalized federated learning (pFL) method for medical visual question answering (VQA) models, addressing privacy reliability challenges in the medical domain. Our method introduces learnable prompts into a Transformer architecture to efficiently train it on diverse medical datasets without massive computational costs. Then we introduce a reliable client VQA model that incorporates Dempster-Shafer evidence theory to quantify uncertainty in predictions, enhancing the model's reliability. Furthermore, we propose a novel inter-client communication mechanism that uses maximum likelihood estimation to balance accuracy and uncertainty, fostering efficient integration of insights across clients. The code will be available soon. 

\keywords{Medical VQA  \and personalized federated learning \and large-scale model.}
\end{abstract}
\section{Introduction}
\label{sec:intro}

Data heterogeneity and privacy protection~\cite{abouelmehdi2018big} are key challenges in the application of deep learning in the healthcare field. Federated Learning (FL)~\cite{mcmahan2017communication}, which allows for learning without centralizing data, is considered a promising solution to this issue. FL is a learning approach that enables the training of distributed models by sharing parameters instead of data.
In the typical problem setting of FL in the medical field~\cite{xu2021federated}, a model is locally trained in each hospital using patient data and sends only the learned model parameters to a central server. The central server aggregates these parameters to create an improved model. This approach allows for improving the model using data from multiple hospitals while protecting patient privacy.
In recent years, research~\cite{10.1145/3501296,sheller2020federated} has been conducted to introduce FL to medical tasks, addressing privacy and data heterogeneity issues. 
Many studies in medical FL focus on simple tasks such as segmentation~\cite{wang2023feddp} and classification~\cite{jiang2023client,zhou2023fedcontrast,10.1007/978-3-031-43895-0_21}, using predefined input and output formats.
Also, most of these studies aim to construct highly versatile models by integrating information from multiple client models.
However, the medical field often requires models that are personalized for specific targets rather than those with broad generalization abilities. It can be clinically more beneficial to fine-tune a model using insights from other models trained on vastly different data rather than relying solely on a single generalized model. 
This is especially important in clinical practice because the same medical images or symptoms often can be interpreted differently for each patient, highlighting the need for personalized models. Hence, FL remains room in the current medical field to tackle more challenging tasks and problem settings.

%

Medical visual question answering (VQA)~\cite{lin2023medical} is a multimodal task capable of handling diverse questions and answers and is considered more challenging than tasks such as segmentation or classification. Implementing a VQA system enables the provision of specialized advice to physicians across different departments and facilitates the delivery of medical advice tailored to specific patients.
In natural images, the proposal of Transformer-based baseline models has significantly improved the accuracy of the VQA task~\cite{vaswani2017attention,devlin2018bert}. However, sufficient consideration has not been given to the application in the medical field, as various clinical constraints need to be taken into account for its implementation.
%
%
%

Personalized Federated Learning (pFL)~\cite{fallah2020personalized} is an approach that focuses on the performance of individual clients rather than the global model of a central server, using protected data to learn personalized models. This approach demonstrates significant compatibility with the medical sector, yet it simultaneously highlights the potential for enhancement in client information aggregation. For example, recent studies~\cite{zhu2024promptbased} based on prompt learning~\cite{wang2022learning,wang2022dualprompt} do not consider the relevance between clients, leading to a degradation of model performance. Assessing the reliability of each client is a crucial issue when considering clinical applications in pFL~\cite{liu2023pre}.

%
%
%

In this paper, we explore the combination of pFL and VQA, tackling a more clinically challenging problem setting. The VQA task has the potential for significant roles in clinical applications with advancements in large language models (LLMs). We propose a novel Transformer model that quantifies the uncertainty of each client and efficiently aggregates beneficial information in this new problem setting.
Specifically, we simulate different departments in a hospital and set up clients for medical images from various organs, each consisting of a VQA model. To reduce the communication burden, we introduce learnable prompts to the Transformer's multi-head attention (MHA) layer, allowing efficient learning of personalized client data distributions.
Furthermore, we propose a Dynamic Likelihood-weighted Uncertainty Calibration (DLUC) process to effectively aggregate information in inter-client communication. This process evaluates the uncertainty of client VQA models using the Dempster-Shafer evidence theory (DST)~\cite{10.5555/3327144.3327239} and dynamically adjusts weights based on this evaluation. This ensures the reliability of the model and supports effective decision-making. 
Extensive qualitative and quantitative experiments on two closed and open-ended medical VQA datasets demonstrate that our method can efficiently aggregate information for personalized clients. 
\section{Methodology}
\begin{figure*}[t]
\centering   
\includegraphics[width=\textwidth]{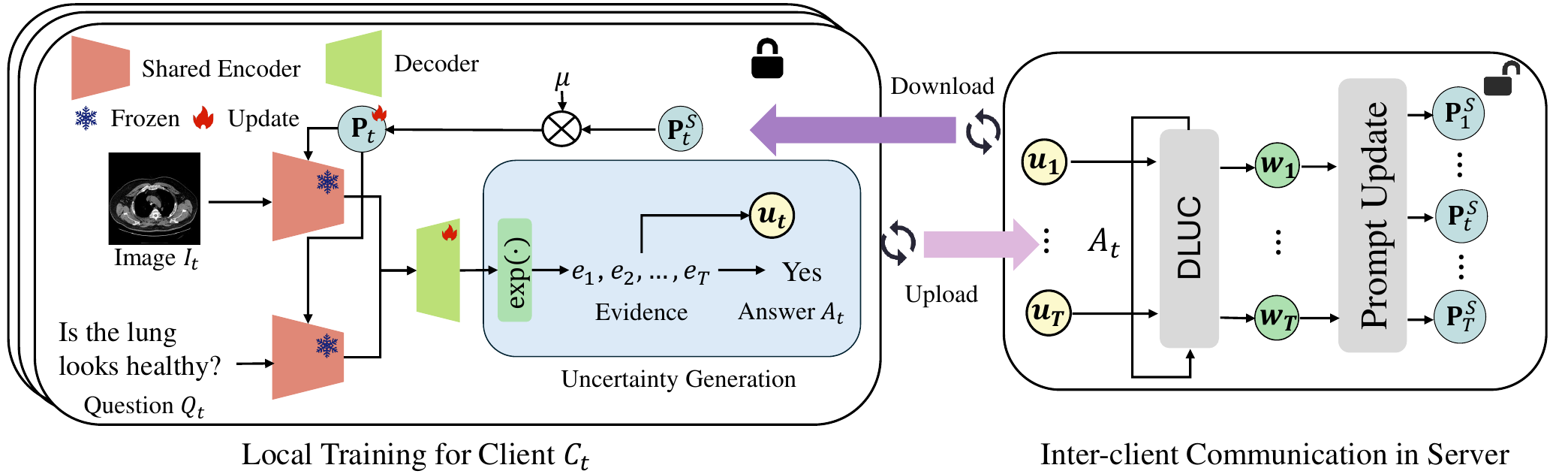}
\caption{Overview of the proposed pFL-VQA framework.}
\label{fig:overview}
\vspace{-1em}
\end{figure*}
\subsection{Problem Formulation}
An overview of our method is shown in Fig.~\ref{fig:overview}. We define $T$ separate clients $\{C_t\}_{t=1}^T$, each of which consists of a VQA model on distributed datasets $\{D_t\}_{t=1}^T$ through the pFL setting.
%
Given a client model $C_t$ with parameter $\theta_t$ and the evaluation metric $Q(\cdot)$, the final optimization objective can be expressed as $\mathop{\arg\max}\limits_{\theta_t} Q(A, U|\theta_t, D_t)$, where $A$ and $U$ denote the generated answer and the uncertainty, respectively. By optimizing this objective, the local training tries to generate $A$ and $U$, which will be used as a criterion of the model's performance evaluation during the next step of the inter-client communication process. Since the reduction in data utilized for training client models inevitably leads to a degradation in performance, the sharing of learned data patterns among clients facilitates a collaborative training process that is mutually advantageous.

Specifically, for client $C_t$, we define the communication method $E$, and the objective is $\mathop{\arg\max}\limits_{E} \sum_{t=1}^T Q(A|E, D_t, \theta_1, ..., \theta_T). \label{Objective-2}$
The objective is to optimize a method that maximizes the information obtained from the other clients for the local training of the client $C_t$.
By optimizing the two mentioned objectives, we can train personalized client models for different medical data distributions without data sharing. This method also maximizes the aggregation of valid information from other clients for each client $C_t$. The local training and inter-client communication processes are described in Sections~\ref{local} and~\ref{inter-client}, respectively.

\subsection{Local Training for Medical VQA Client Model} \label{local}
%
For client $C_t$, the dataset $D_t$ comprises $n$ instances that consist of an input question $Q_t$, an image $I_t$, and the corresponding ground truth answer $A_t$. 
We utilize Transformer-based encoders $E_I$ and $E_Q$ to extract features of the input image and question separately. Subsequently, an answering decoder $D$ generates the predicted answer.

In the proposed method, we add learnable prompts $\mathbf{p}\in \mathbbm{R}^{d\times m}$ ($d$ is the length of the prompt, $m$ is the hidden dimension of the Transformer) to the multi-head attention (MHA) layers of the encoders.
Specifically, we use Prefix-Tuning~\cite{li2021prefix} splitting $\mathbf{p}$ into $\{ \mathbf{p}^K, \mathbf{p}^V\} \in \mathbbm{R}^{d/2 \times m}$, which are introduced into the MHA layer as follows:
\begin{equation}
    Attention(\mathbf{Q}, \hat{\mathbf{K}}, \hat{\mathbf{V}}) = softmax(\frac{\mathbf{Q}\hat{\mathbf{K}}^{T}}{\sqrt{m}})\hat{\mathbf{V}};\    \hat{\mathbf{K}} = \left[\mathbf{p}^k; \mathbf{K}\right],
    \hat{\mathbf{V}} = \left[\mathbf{p}^v; \mathbf{V}\right],
\end{equation}
where $\mathbf{Q}$, $\mathbf{K}$, and $\mathbf{V}$ are the original inputs of the MHA layer.

Conventional VQA models output the prediction for the answer probability directly from the hidden state of the decoder output via a softmax activation function. 
In order to capture the uncertainty for evaluating the model performance, based on the DST theory, we consider the probability of answers to obey a Dirichlet distribution with parameter $\mathbf{\alpha}_j = \{\alpha_{j}\}_{j=1}^J$ ($J$ is the length of the answer list). 
Specifically, we define the output hidden states of the decoder $D$ as the evidence $\mathbf{e} = \{e_{j}\}_{j=1}^J$ of each candidate's answer. 
Since the evidence is non-negative, in order to get the evidence, we replace the final softmax layer of the conventional VQA models with the activation layer $exp(\cdot)$, and $\mathbf{e}$ can be calculated as $e_{j} = exp(f(D_t |\theta_t))$.
The Dirichlet parameter $\mathbf{\alpha} = \{\alpha_{j}\}_{t=1}^T$ can be calculated by evidence as $\alpha_{j} = e_{j} + 1$.
Based on $\mathbf{\alpha}$, the belief mass $\mathbf{b} = \{b_{j}\}_{j=1}^J$ of the model for each candidate's answer can be defined as $b_{j} = \frac{e_{j}}{S}$, where $S = \sum_{j=1}^J\alpha_{j}$. 
Unlike traditional probability distributions, the sum of $\mathbf{b}$ does not equal 1, and the difference from 1 is the uncertainty $u$ of the model, which can be calculated as $u = 1 - \sum_{j=1}^Jb_{j}; u > 0, b_{j} > 0$.

To assign more evidence to the correct answer, we employ the following Dirichlet term:
\begin{equation}
    Dirichlet(\theta_t) = \int \frac{1}{B(\mathbf{\alpha}_j)}\prod_{j=1}^Jp_{j}^{\alpha_{j}-1}d\mathbf{p}_j, 
\end{equation}
where $B(\cdot)$ is the beta function. The Dirichlet term is introduced into the cross-entropy loss function according to obtain $\mathcal{L}^{unc}(\cdot)$ as follows:
\begin{equation}
        \mathcal{L}^{unc}(\theta_t) = \sum_{j=1}^J y_{j}(\psi(S_j)-\psi(\alpha_{j})), \label{loss_unc}
\end{equation}
where $y_{j}$ is the one-hot vector of the ground truth answer, and $\psi(\cdot)$ is the digamma function. Note that Eq.~(\ref{loss_unc}) aims to allocate the sum of all evidence generated by predictions as much as possible to the correct answers, providing positive feedback. However, since the above loss does not ensure that incorrect labels yield less evidence, the KL term is introduced to minimize the evidence for incorrect labels as much as possible to 0:
\begin{align}
        KL[D(\mathbf{p}_j|\hat{\mathbf{\alpha}}_j)||D(\mathbf{p}_j|\mathbf{1})] &= log(\frac{\Gamma(\sum_{j=1}^J \hat{\alpha}_{j})}{\Gamma(J)\prod_{j=1}^J\Gamma(\hat{\alpha}_{j})}) \\
        &+ \sum_{j=1}^J(\hat{\alpha}_{j}-1)[log(\hat{\alpha}_{j})- log(\sum_{i=1}^J \hat{\alpha}_{j})]
\end{align}
Here, $\mathbf{1}$ is a parameter vector of $K$ ones, and $\hat{\mathbf{\alpha}}_j=\mathbf{y}_j+(1-\mathbf{y}_j)\odot \alpha_i$ is the Dirichlet parameter to prevent penalties for correct evidence (when $y_j= 1$, $\hat{\mathbf{\alpha}}_j = 1$ and the loss becomes 0). Therefore, the final loss function can be calculated as follows:
\begin{equation}
\mathcal{L}(\theta) = \mathcal{L}^{unc}(\theta) + \lambda KL[D(\mathbf{p}_j|\hat{\alpha}_j)||D(\mathbf{p}_j|\mathbf{1})],
\end{equation}
where $\lambda$ is a balance parameter. The client model can output the answer text and the uncertainty $U_n$ by training the model using the proposed loss function. In the inter-client communication process, uncertainty evaluates the model's performance.
\vspace{-1em}
\subsection{DLUC-based Inter-client Communication Process} \label{inter-client}
%
%
We define a dynamic weight $w_t$ for client $C_t$, and the prompt $\mathbf{p}_t$ is added with the weighted sum of the prompts of the other clients $\mathbf{p}_t^S$.
Given the data $D_t$ and the clients' parameters $\theta_t$, the conditional expectation, as a lower bound on the true likelihood, is computed as
\begin{equation}
    \mathbbm{E}\{log Pr[D_t|\theta_t]\} = \sum_{t=1}^T log Pr(A_{t}|D_t, \theta_t, \mathbf{p}_t + \mu\mathbf{p}_t^S), \label{likelihood}
\end{equation}
where $\mathbf{p}_{t}^S = \sum_{k=1}^{T,k\neq t} w_{k}\mathbf{p}_{k}$, $\mu$ is an aggregation rate, and the $w_t$ is the weight optimized by the EM method $\mathbbm{E}$.
Since the prompts represent the client data schema, we use a neural network to optimize the objective of Eq.~(\ref{likelihood}). We take the uncertainty $u$ as the network input and the weights $w$ as the output. During training, we set up local training step $step_{l}$ and inter-client communication $step_{c}$ separately, representing the number of iterations each is executed, and cross these two processes. This DLUC aims to find the best configuration of weights to maximize the total likelihood of the model for all client data. This optimization allows the client to aggregate the maximum amount of information from the server while avoiding the interference of invalid information. 
\vspace{-1em}
\section{Experiments}
We are motivated by the excellent performance of transformers on multimodal tasks in the medical domain. Considering the wide range of VQA task applications, we selected medical images and QA pairs from VQA-RAD~\cite{lau2018dataset} and Slake~\cite{liu2021slake} datasets. These datasets were further partitioned into sub-datasets based on the anatomical focus of the images. With reference to the setup of different departments in the actual clinic, we split Slake into ``Lung'' (Client1), ``Abdomen'' (Client2), ``Brain'' (Client3), and ``Other'' (Client4), VQA-RAD into  ``Abdomen'' (Client1), ``Chest'' (Client2), ``Head'' (Client3), respectively. Details of the dataset can be found in Table 1 of Supplementary.
To mitigate the impact of accuracy fluctuations observed among some clients, attributed to the limited size of their test datasets, we determined the final accuracy based on the average results and the variance from the last round of local training.
Furthermore, we also tested the method's performance when the number of clients spikes using the large-scale dataset PMC-VQA~\cite{zhang2023pmcvqa}.
\begin{table}[t]
\scriptsize
\centering
\caption{Quantitative results of accuracy comparison with previous medical VQA models. The Enc. and Upd. denote the number of parameters in the baseline and parameters that need to be updated during training, respectively.}
\label{Tabel:Acc ALL}
\begin{tabular}{ccccclcccccccccc}
\hline
\multirow{2}{*}{Method} & \multicolumn{8}{c}{VQA accuracy (\%)}& \multicolumn{2}{c}{Param. (M)} \\ \cline{2-11} 
            &\multicolumn{4}{c}{Slake} && \multicolumn{3}{c}{VQA-RAD} & Enc.        & Upd.         \\ \cline{1-5} \cline{7-11} 

PMC-VQA~\cite{zhang2023pmcvqa}& \multicolumn{4}{c}{82.5}&& \multicolumn{3}{c}{86.8} &7000         & 7000 \\

BioMedGPT~\cite{zhang2023biomedgpt}& \multicolumn{4}{c}{82.5}&& \multicolumn{3}{c}{81.3} &1500         & 1500 \\

LoRA~\cite{SonsbeekOpen2023}&\multicolumn{4}{c}{82.1}&& \multicolumn{3}{c}{-}    &1500         & 15   \\
M2I2~\cite{li2023self}&\multicolumn{4}{c}{83.2}&& \multicolumn{3}{c}{70.8} &600          &600    \\
BioMedCLIP~\cite{zhang2024biomedclip}&\multicolumn{4}{c}{86.7}&& \multicolumn{3}{c}{79.8} &150          & 340   \\
\rowcolor{gray!25}
MEVF-BAN~\cite{10.1007/978-3-030-32251-9_57}& \multicolumn{4}{c}{75.4}&\cellcolor{white}& \multicolumn{3}{c}{75.1} &50           & 50 \\
\rowcolor{gray!25}
VGG+SAN~\cite{liu2021slake}& \multicolumn{4}{c}{75.4}&\cellcolor{white}& \multicolumn{3}{c}{74.0} &50           & 50  \\\cline{1-5} \cline{7-11}\cline{1-5} \cline{7-11} 
            & Client1 & Client2 & Client3 & Client4 &&Client1 & Client2 & Client3   && \\ \cline{1-5} \cline{7-11} 
\rowcolor{gray!25}
PMVQA~\cite{zhu2024promptbased}&84.4&72.3&76.9&81.4&\cellcolor{white}&68.9&64.7&74.2&63&0.3  \\ 
\rowcolor{gray!25}
PM    &$81.7_{\pm0.5}$&$\mathbf{74.7}_{\pm0.8}$&$\mathbf{81.3}_{\pm0.3}$&$78.2_{\pm1.0}$&\cellcolor{white}&$\mathbf{72.1}_{\pm1.2}$&$\mathbf{65.1}_{\pm0.5}$&$\mathbf{81.2}_{\pm0.8}$&63&\textbf{0.01}\\ \hline
\end{tabular}
\vspace{-2em}
\end{table}
\begin{table}[t]
\scriptsize
\centering
\caption{Comparison of VQA accuracy (\%) across several common baseline networks before and after applying the proposed method.}
\begin{tabular}{cccccclccc}
\hline
\multirow{2}{*}{Baseline} &\multirow{2}{*}{Param.}&\multicolumn{4}{c}{Slake} && \multicolumn{3}{c}{VQA-RAD} \\ \cline{3-6}\cline{8-10}
                         && Lung              & Abdomen            & Brain              & Others & & Chest    & Abdomen    & Head    \\ \hline
VIT-B/32~\cite{radford2021learning}&63M& $-1.2_{\pm1.0}$&$\mathbf{+1.8}_{\pm0.8}$ & $\mathbf{+0.9}_{\pm0.2}$ & $\mathbf{+4.7}_{\pm1.7}$&&$\mathbf{+2.8}_{\pm1.6}$&$\mathbf{+2.1}_{\pm0.1}$&$\mathbf{+0.6}_{\pm0.2}$\\

PubMedCLIP~\cite{eslami2021does}&63M&$-0.3_{\pm0.5}$&$\mathbf{+0.3}_{\pm0.9}$&$\mathbf{+6.8}_{\pm0.9}$&$\mathbf{+4.7}_{\pm0.4}$&&$\mathbf{+0.2}_{\pm1.2}$&$-2.3_{\pm1.4}$&$\mathbf{+2.7}_{\pm1}$\\    
VIT-L/14~\cite{radford2021learning}&300M& $-0.7_{\pm1.0}$ & $\mathbf{+0.3}_{\pm1.2}$ & $\mathbf{+0.2}_{\pm0.8}$ & $\mathbf{+3.0}_{\pm0.7}$ & &$\mathbf{+2.8}_{\pm2.1}$&$\mathbf{+2.4}_{\pm1.4}$&$-1.52_{\pm2.5}$ \\
ViT-L/14@336px~\cite{radford2021learning}&300M& $-1.2_{\pm0.2}$ &$-2.7_{\pm0.6}$&$\mathbf{+2.3}_{\pm0.4}$&$\mathbf{+9.5}_{\pm2.9}$&&$\mathbf{+1.1}_{\pm0.8}$&$\mathbf{+2.0}_{\pm1.4}$&$\mathbf{-0.4}_{\pm2.5}$\\\hline
\end{tabular}
\label{Table: multibaseline}
\end{table}
\begin{table}[t]
\scriptsize
\centering
\caption{Ablation study on Slake.}
\label{Tabel:Slake abl}
\begin{tabular}{ccccclcccccc}
\hline
    & \multicolumn{4}{c}{CLOSED}  && \multicolumn{4}{c}{OPEN}   \\ \cline{2-5}\cline{7-10}
    & Client1 & Client2 & Client3 & Client4  && Client1 & Client2 & Client3 & Client4   & \\ \cline{1-5}\cline{7-10}
PM w/o client &\multicolumn{4}{c}{$77.8_{\pm0.2}$}&&\multicolumn{4}{c}{$58.2_{\pm0.4}$}&\\
PM w/o server &$\mathbf{83.7}_{\pm0.4}$&$71.3_{\pm0.7}$&$80.8_{\pm0.6}$&$74.9_{\pm1.6}$&&$59.5_{\pm0.2}$&$\mathbf{56.3}_{\pm0.8}$&$51.1_{\pm0.9}$&$43.8_{\pm0.8}$\\ 
PM w/o DLUC &$79.9_{\pm0.4}$&$74.8_{\pm0.6}$&$80.9_{\pm0.4}$&$78.4_{\pm0.4}$&&$65.9_{\pm0.1}$&$55.9_{\pm0.2}$&$63.7_{\pm0.5}$&$61.1_{\pm0.8}$   \\ 
PM  &$81.7_{\pm0.5}$&$74.7_{\pm0.8}$&$\mathbf{81.3}_{\pm0.3}$&$\mathbf{78.9}_{\pm1.0}$&&$\mathbf{66.0}_{\pm0.3}$&$56.1_{\pm0.3}$&$\mathbf{63.8}_{\pm0.2}$&$\mathbf{61.7}_{\pm0.8}$  \\ \hline
\end{tabular}
\vspace{-2.5em}
\label{Table:abSlake}
\end{table}
\begin{table}[t]
\scriptsize
\centering
\caption{Ablation study on VQA-RAD.}
\label{Tabel:VQA-RAD abl}
\begin{tabular}{cccclccccccc}
\hline
    & \multicolumn{3}{c}{CLOSED}  && \multicolumn{3}{c}{OPEN}   \\ \cline{2-4}\cline{6-8}
    & Client1 & Client2 & Client3  && Client1 & Client2 & Client3   & \\ \cline{1-4}\cline{6-8}
PM w/o client & \multicolumn{3}{c}{$67.5_{\pm0.9}$}&&\multicolumn{3}{c}{$42.6_{\pm0.4}$}&       &\\
PM w/o server &$68.8_{\pm0.7}$&$62.9_{\pm1.6}$&$80.8_{\pm1.4}$&&$44.2_{\pm0.2}$&$38.2_{\pm0.3}$&$34.5_{\pm0.1}$&  \\ 
PM w/o DLUC &$71.0_{\pm0.5}$&$61.8_{\pm0.6}$&$76.5_{\pm0.8}$&&$47.7_{\pm0.2}$&$39.9_{\pm0.7}$&$\mathbf{35.3}_{\pm0.2}$&   \\ 
PM  &$\mathbf{72.1}_{\pm1.2}$&$65.1_{\pm0.9}$&$\mathbf{81.2}_{\pm0.8}$&&$\mathbf{47.9}_{\pm0.2}$&$40.3_{\pm0.3}$&$34.3_{\pm0.3}$&   \\ \hline
\end{tabular}
\label{Table:abVQARAD}
\vspace{-2em}
\end{table}

For specific modeling details, we employed the pre-trained Contrastive Language–Image Pre-training (CLIP)~\cite{radford2021learning} model as the image and text encoders $E_I$ and $E_Q$. All clients share a CLIP with a fixed weight, and only the client's prompts are updated during training.
In our experiments, we got the best results when we added prompts in the first four blocks of the Transformer and set the length of the prompts to 24 and 30 for the Slake and VQA-RAD, respectively. The reason is that the feature in different organs focus more on patterns in smaller regions, and the Transformer's beginning few blocks outputs tend to capture more localized information.
The learning rate of the prompts is initially 0.001, multiplied by 0.5 after each $Step_{l}$, i.e., one complete local update.
The answer layer is implemented as a 2-layer MLP with a hidden size 512 and a dropout rate of 0.2. We used a fully connected layer for the DLUC, whose learning rate is initially 0.01 and multiplied by 0.1 after each complete inter-client communication.

\noindent \textbf{Comparison Methods:} Given the complexity and the substantial size of the baseline model employed, along with the intricacies of the multimodal task setup, direct comparison with previous FL methods is not feasible. Considering that we aim to enhance the personalized performance of medical VQA, we compare the proposed with several state-of-the-art medical VQA models regarding accuracy and model parameters that need to be updated. Specifically, we employed PMC-VQA~\cite{zhang2023pmcvqa}, BioMedGPT~\cite{zhang2023biomedgpt}, M2I2~\cite{li2023self}, BioMedCLIP~\cite{zhang2024biomedclip}, MEVF-BAN~\cite{10.1007/978-3-030-32251-9_57}, VGG+SAN~\cite{liu2021slake}, LoRA~\cite{SonsbeekOpen2023} and PMVQA~\cite{zhu2024promptbased}. 

As the results shown in Table~\ref{Tabel:Acc ALL}, compared to the PMVQA~\cite{zhu2024promptbased} with the same task setup, we averaged a 3.5\% accuracy improvement on most clients. Compared to other locally trained models, our method achieves higher precision when the number of baseline parameters is close. However, there is still a gap when comparing methods using more parameter baselines. It is worth noting parameters that need to be updated by our method are \textit{0.01\% of the Vision Transformer-based model} and \textit{0.00001\% of the current SOTA model}. 

To demonstrate that our methodology can be effectively extended to all transformer structure networks, we tested the performance of our method on multiple baselines. As shown in Table~\ref{Table: multibaseline}, our method can effectively improve the personalization performance of multimodal transformers, and we achieved an \textit{average of 3\% improvement} on most clients. Further analysis of the differences in weights and performance across clients is in the supplementary.

\noindent \textbf{Ablation studies.} We designed: trained on the whole data without setting up the client, setting up the client but without parameter sharing in sever, and without DLUC for information aggregation (equal weights). As shown in Table~\ref{Table:abSlake} and Table~\ref{Table:abVQARAD}, the results demonstrate that DLUC can effectively aggregate clients' information to improve the performance of each client. 

\begin{figure}[t]
  \centering
    \begin{minipage}[b]{0.35\linewidth}
            \includegraphics[width=\linewidth]{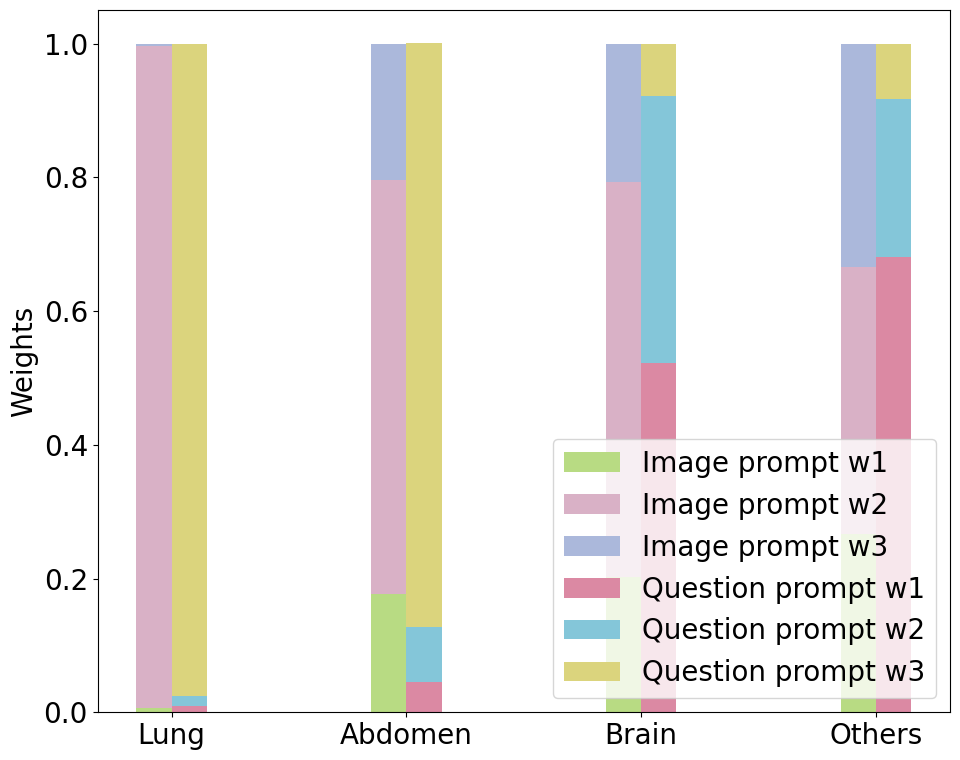}
    \end{minipage}
    \begin{minipage}[b]{0.35\linewidth}
            \includegraphics[width=\linewidth]{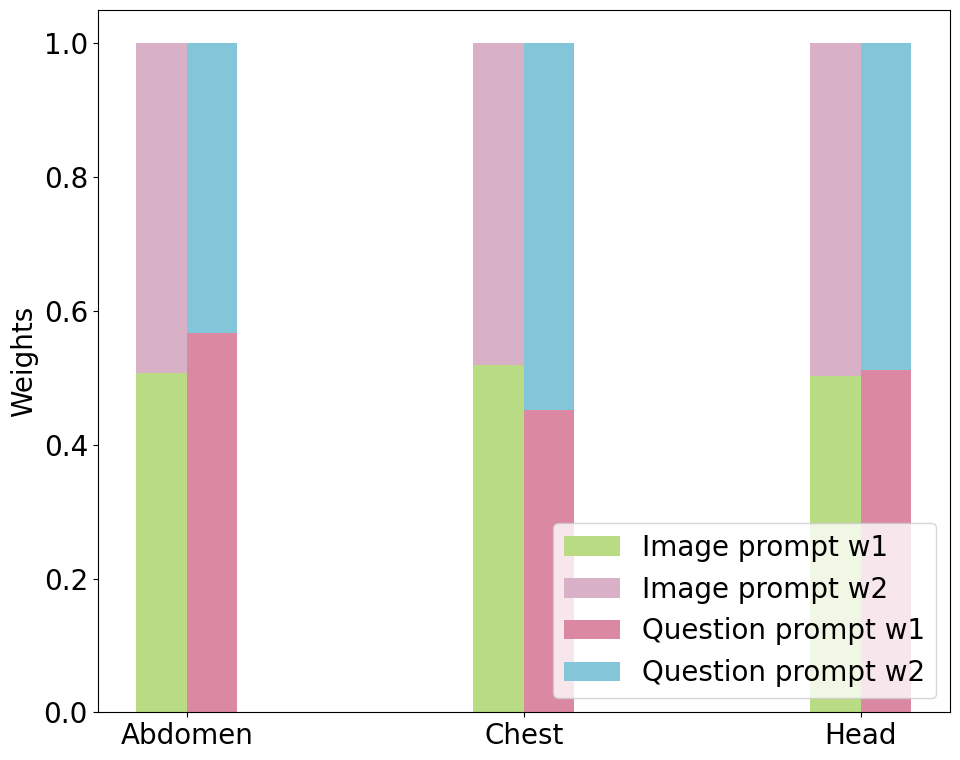}
    \end{minipage}
      \caption{Weights generated by DLUC for clients during inter-client communication.} \label{Fig:Weights}
\end{figure}

\noindent \textbf{Analysis of inter-client weights.} Fig.~\ref{Fig:Weights} illustrates the weights generated by DLUC during inter-client communication for each client, and the results show significant differences in the dependency between clients. For example, the image prompt of ``Lung"  is highly dependent on ``Abdomen" information because the images of these client in the Slake are tomographic CT scans from immediate area that have similar features. Furthermore, the images from the three clients of VQA-RAD are obtained from CT, X-ray, and MRI, so there is no significant dependence on the results due to the large modality gap. \textit{The results show that our generation of weights is clinically meaningful.}

\begin{figure}
  \centering
  \begin{minipage}[b]{0.55\linewidth}
    \begin{minipage}[b]{0.49\linewidth}
            \includegraphics[width=\linewidth]{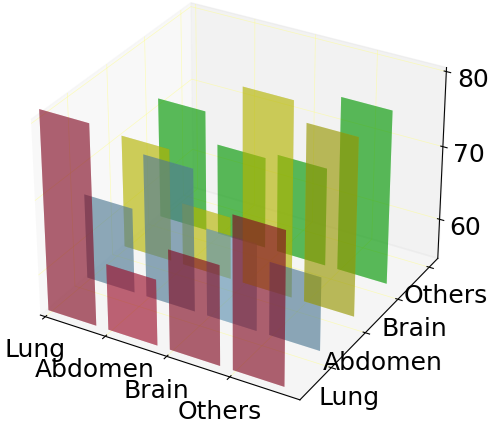}
    \end{minipage}
    \begin{minipage}[b]{0.49\linewidth}
            \includegraphics[width=\linewidth]{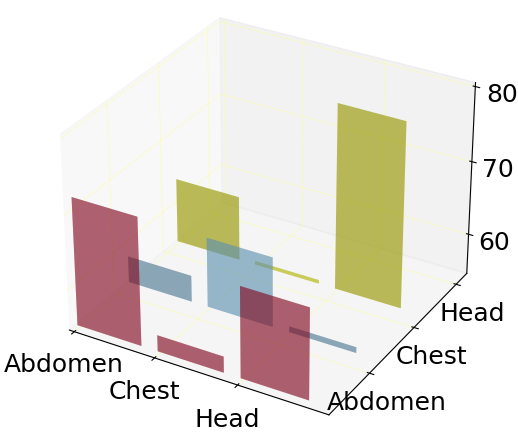}
    \end{minipage}
    \caption{The cross-evaluation results.} \label{Fig:cross-eva}
    
  \end{minipage}
    \begin{minipage}[b]{0.44\linewidth}
            \includegraphics[width=\linewidth]{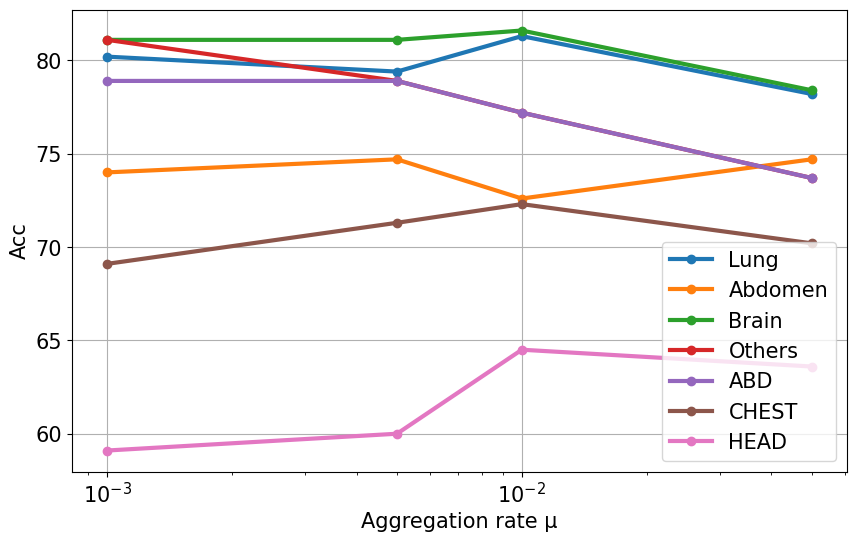}
    \caption{Comparison on different $\mu$.}  \label{Fig:Aggregation}
    \end{minipage}
\vspace{-2.5em}
\end{figure}

\noindent \textbf{Cross evaluation between prompts.} To further demonstrate that prompts can effectively personalize the model rather than learn generalizability information, we performed accuracy tests for each client using prompts from the other clients. As shown in Fig.~\ref{Fig:cross-eva}, there is a significant drop in accuracy when the prompts are exchanged, proving that the prompts learn information about different data distribution.

\noindent \textbf{Impact of hyperparameters.} The model's performance is affected by three primary parameters: the local training step $step_{l}$, the DLUC step in inter-client communication $step_{c}$, and the aggregation rate $\mu$. The comparison results on step are shown in Fig.~1 in Supplementary. The results shows that higher $step_{l}$ tends to perform better, and performances on higher $step_{c}$ are better when the $step_{l}$ is low. This is because when there is less local training, the model cannot fully adapt to the local data, and optimizing the aggregation weights by DLUC can integrate information from different clients more effectively. As for $\mu$, both too high and too low result in deterioration of the results due to excessive or insufficient influence of external information as shown in Fig.~\ref{Fig:Aggregation}.

\noindent \textbf{Extreme Tests:} Since each of our clients essentially retains only learnable prompts with only 0.01M parameters, our method can set up a tremendous number of clients, which was absolutely difficult to achieve with the previous pFL method. We experimented with setting up 500 clients on the PMCVQA dataset, consumed only 13G of VRAM, and experimental results show an average 1.8\% accuracy improvement on 56\% of the clients.
\vspace{-1em}
\section{Conclusion}
\vspace{-1em}
We present a prompt-based, reliable pFL method for medical VQA. Our method creates clients for heterogeneous medical data and achieves aggregation of information through the novel DLUC. In the DLUC process, we calculate the uncertainty of each client through DST theory and generate aggregation weights based on maximum likelihood estimation through uncertainty. The experimental results demonstrate that our method can effectively improve the performance of the transformer-based client model with minimal overhead. Then, the computation of our method is independent of the size of the baseline network and can be applied in personalized learning with any transformer of any size. In future work, we will introduce the method to personalize larger-scale models.
%
%
%
\bibliographystyle{splncs04}
\bibliography{main}
%




\end{document}